\documentclass[sigconf]{acmart}
\usepackage{cleveref}
\usepackage{booktabs}
\usepackage{multirow}
\usepackage{makecell}
\usepackage{tikz}
\usepackage{xcolor}
\usepackage[normalem]{ulem}
\definecolor{revcolor}{rgb}{1,0,0}

\definecolor{revdelcolor}{rgb}{0,0,1}

\newcommand{\redcircle}[1]{%
  \,
  \begin{tikzpicture}[baseline=(char.base)]
    \node[circle, fill=red, text=white, inner sep=1.5pt, font=\bfseries] (char) {#1};
  \end{tikzpicture}%
  \,
}
\newcommand{\revdone}[1]{#1}
\newcommand{\revdeleted}[1]{}

\AtBeginDocument{%
  }

\setcopyright{acmlicensed}
\copyrightyear{2026}
\acmYear{2026}
\setcopyright{cc}
\setcctype{by}
\acmConference[SIGSPATIAL '26]{The 34th ACM International Conference on Advances in Geographic Information Systems}{November 03--06, 2026}{Riverside, CA, USA}
\acmBooktitle{The 34th ACM International Conference on Advances in Geographic Information Systems (SIGSPATIAL '26), November 03--06, 2026, Riverside, CA, USA}
\acmDOI{10.1145/3841645.3843316}
\acmISBN{979-8-4007-2950-8/2026/11}

\begin{document}

\title{Applying foundation model embeddings towards urban livability evaluation}

\author{Ayush Khot}
\email{akhot2@illinois.edu}
\affiliation{%
  \institution{Siebel School of Computing and Data Science, University of Illinois at Urbana-Champaign}
  \city{Urbana}
  \state{IL}
  \country{USA}
}

\author{Wen Zhou}
\email{wz53@illinois.edu}
\affiliation{%
  \institution{Department of Geography and Geographic Information Science, University of Illinois at Urbana-Champaign}
  \city{Urbana}
  \state{IL}
  \country{USA}
}

\author{Shaowen Wang}
\correspondingauthor
\email{shaowen@illinois.edu}
\affiliation{%
  \institution{Department of Geography and Geographic Information Science \& Siebel School of Computing and Data Science, University of Illinois at Urbana-Champaign}
  \city{Urbana}
  \state{IL}
  \country{USA}
}

\renewcommand{\shortauthors}{Khot et al.}

\begin{abstract}
  While accurate measurement of socioeconomic indicators remains challenging in data-scarce regions, which limits policy interventions and resource allocation, high-resolution geospatial data is widely available and can contain information on various livability statistics. We investigate which physical features are encoded within foundation model embeddings, such as AlphaEarth, AnySat, and TerraMind, and provide a systematic framework for identifying the most predictive geospatial indicators. By analyzing how different types of geospatial data influence urban livability predictions, our approach enables researchers to prioritize the most informative features for their specific applications. Additionally, we demonstrate how to leverage foundation model embeddings to enhance prediction performance for these outcomes. This work contributes a principled methodology for extracting actionable information from satellite imagery while accounting for complex spatial dependencies, with applications in predicting urban livability in regions with limited observation data.
\end{abstract}

\ccsdesc[500]{Computing methodologies~Scene understanding}
\ccsdesc[500]{Information systems~Geographic information systems}
\ccsdesc[300]{Computing methodologies~Transfer learning}

\keywords{AlphaEarth, TerraMind, AnySat, Geospatial Embeddings, Urban livability, Multimodal deep learning, Satellite images, Digital surface model, Nighttime light remote sensing, Textual information}


\maketitle

\section{Introduction}
\label{sec:intro}

Urbanization is reshaping the modern world. Nearly 45\% of the world's 8.2 billion people currently live in cities, and this share is projected to reach 68\% of the world's 9.7 billion people by 2050, with two-thirds of global growth projected to occur in cities within the next 25 years. 
As such, it is important to understand and improve the quality of life within cities. Although well-being is subjective, the Sustainable Development Goal 11 provides a shared framework, calling for cities and settlements to be inclusive, safe, resilient, and sustainable.

Previous research on assessing urban livability has focused on statistical methods to estimate livability metrics~\cite{gis-livability-eval, shanghai, vienna}. However, these methods are not generalizable as indicators are selected manually. The authors of \cite{zhou_livability} address this gap by using a variety of geospatial data for deep learning-based livability evaluation. 
These data sources are collected and processed independently, requiring significant effort to integrate. Furthermore, potentially valuable datasets may have been overlooked or excluded due to collection costs, time constraints, or subjective judgement.
Geospatial embeddings offer a promising avenue to address these drawbacks.
These foundational embeddings assimilate a diverse range of data sources, including optical, radar, environmental, and textual information into real-numbered vectors. This could reduce data engineering overhead while potentially improving livability predictions through unifying data modalities. We specifically examine AlphaEarth Foundations~\cite{alphaearth}, TerraMind~\cite{terramind}, and AnySat~\cite{anysat} to understand their impact on urban livability evaluation. We build on the Transformer from \cite{zhou_livability} to include additional embeddings and explore how these representations interact with different data modalities.
\revdone{
The contribution of this paper lies not in extending the model to incorporate embeddings, but rather in analyzing how different geospatial embeddings, both individually and in combination, influence livability prediction. We also conduct interpretability analyses to better understand how these embeddings affect the model's internal representations and decision-making processes.}

\section{Related Work}
\label{sec:related}

\subsection{Urban Livability}
Previous research on assessing livability has focused on using statistical methods with chosen indicators. For example, 20 objective indicators were identified by statistical, remote sensing, and spatial data to assess the livability of urban areas using an analytic hierarchy process method~\cite{gis-livability-eval}. The authors of ~\cite{shanghai} have focused on making livability multi-dimensional instead of over-emphasizing on economic aspects by relying on five major categories: education, medical services, recreation, transportation services, and living services. Recent work has used government data~\cite{opendataaustria2024} and OpenStreetMap~\cite{OpenStreetMap} to introduce a foundational livability indicator averaged across two domains: accessibility and economy.
While these approaches offer valuable insights into livability, they often lack consistency between regions or fail to integrate diverse data sources. 
The Leefbaarometer project (LBM)~\cite{lbm} initiated by the Dutch government addresses these limitations by integrating a broad range of livability-influencing indicators into nationally consistent evaluation results, identifying 47 environmental characteristics linked to how much residents enjoy and value living in their areas. 
Although many geospatial datasets hold information relevant to livability~\cite{bigdata}, only 8.8\% of the 68 studies reviewed in \cite{indicators} make use of them.
The authors of \cite{zhou_livability} address this underutilization through a transformer-based deep learning model to evaluate urban livability. They use a wide variety of large-volume data sources, which can be difficult to collect, preprocess, integrate, and train. Geospatial embeddings can serve as a gateway towards bridging this gap by offering precomputed, spatially grounded representations that consolidate diverse data sources without requiring task-specific collection and processing pipelines.

\subsection{Geospatial Embeddings}
In this work, we examine three different foundational embeddings: AlphaEarth, AnySat, and TerraMind.
AlphaEarth Foundations~\cite{alphaearth} is an embedding field model that unifies a wide variety of Earth observation data. They use a transformer-based architecture to fuse the data sources into unit-normalized 64-dimensional vectors, generated at a 10-meter spatial resolution. These are already being used to enhance geospatial tasks, including data generation~\cite{alphadatageneration}, retrieval-augmented generation~\cite{rag}, and height mapping~\cite{heightmapping}. AnySat~\cite{anysat} is a self-supervised geospatial model based on joint embedding predictive architecture and scale-adaptive spatial encoders. They use modalities with different resolutions to create 1536-dimensional vectors with a spatial resolution of 10m. The authors demonstrated its use on tasks like land cover mapping, deforestation detection, and flood segmentation~\cite{anysat}. Finally, TerraMind~\cite{terramind} is another multimodal foundation model that introduces a dual-scale encoder-decoder architecture combining token-level and pixel-level representations across many modalities. These 384-dimensional embeddings are offered at 160m spatial resolution, and they have been applied towards wildfire mapping~\cite{terramind-wildfire} and flood susceptibility mapping~\cite{terramind-flood}.

\section{Data}
\label{sec:data}

\begin{figure}[tp]
    \centering
    \includegraphics[width=\linewidth]{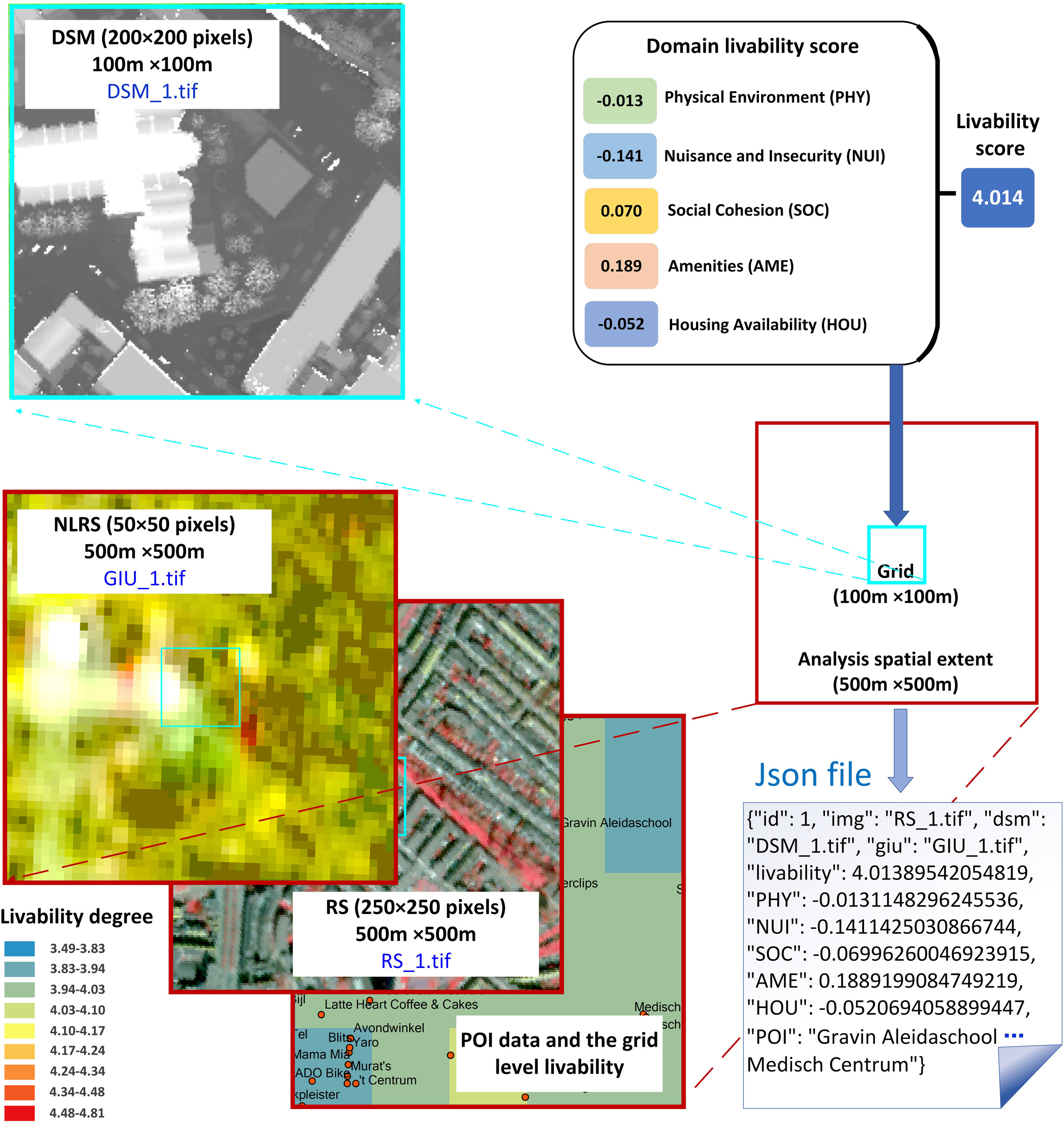}
    \caption{Overview of the dataset generation pipeline~\cite{zhou_livability}. Remote sensing (RS), digital surface model (DSM), nightlight remote sensing (NLRS), and point-of-interest (POI) inputs are extracted at two spatial scales: DSM and livability scores cover the 100m $\times$ 100m target grid, while RS, NLRS, and POI span a broader 500m $\times$ 500m context window.}
    \Description{Diagram showing the images of the inputs at two different spatial scales along with example numbers for the livability score. This is all stored in a json file.}
    \label{fig:lbm}
\end{figure}

The Leefbaarometer (LBM) project~\cite{lbm} version 3.0 maintains national-level urban livability measurements across various cities in the Netherlands. They provide a wide range of spatial scales: grid, neighborhood, district, postcode, and municipality. To maintain consistency with raster data, we use the LBM's smallest spatial unit, the 100m $\times$ 100m grid. The overall livability is split up into five dimensions: physical environment (PHY), housing availability (HOU), amenities (AME), social cohesion (SOC), and nuisance and insecurity (NUI). The range of livability scores is from 3.3464 to 4.853, PHY from -0.142 to 0.118, NUI from -0.514 to 0.111, SOC from -0.151 to 0.095, HOU from -0.344 to 0.164, and AME from -0.080 to 0.695.

\revdone{In the LBM project, the authors aggregate 47 environmental characteristics into these five domains~\cite{lbm, zhou_livability}. PHY is the broadest, spanning proximity to infrastructure such as major roads and high-voltage lines, natural amenities like green space and water, land-use mix, and risk factors including flooding, heat stress, and air quality. HOU is narrower, covering residential areas, housing vacancy, and the height, type, and age of buildings. AME reflects distance to and density of educational, healthcare, cultural, and retail facilities, along with transit and job accessibility. SOC combines perceived social cohesion with residential turnover, population density, household growth or decline, and age diversity. NUI captures reported violent crime, vandalism, and disorderly conduct, alongside residents' perceived disorder and fear of crime.
}

Following \cite{zhou_livability}, we select the following input data: remote sensing (RS) images, digital surface model (DSM), nightlight remote sensing (NLRS) images and point-of-interest (POI) data. These modalities often have information that capture livability and its many domains. For example, RS data can infer car density information~\cite{cardensity} (PHY), NLRS can infer population density~\cite{nlrs-population} (SOC) and burglary~\cite{nlrs-burglary} (NUI), DSM can infer building height (HOU), and POI data can infer facility use (AME). 
No fixed set of modalities can exhaustively capture all factors that influence livability.
As such, we also consider AlphaEarth, AnySat, and TerraMind embeddings to understand whether they can capture urban livability statistics. All data sources correspond to the year 2020.
The RS images are sourced from SuperView multispectral satellite images at a 2m spatial resolution. The DSM is provided by the Dutch public geo-services platform (PDOK) with a 0.5m spatial resolution. The NLRS images are SDGSAT-1 Glimmer images at a 10m spatial resolution. All of the embeddings are accessible through the \texttt{rs-embed} tool ~\cite{rs-embed}.
\footnote{\url{https://github.com/cybergis/rs-embed}}

\Cref{fig:lbm} illustrates how the dataset was constructed. We consider a buffer of 200m outside the evaluation grid cell, creating a total spatial extent of 500m. The DSM was confined within the grid while the RS, NLRS, POI, and embeddings included the buffer. As such, the pixel dimensions for RS, DSM, NLRS, AlphaEarth, and AnySat are respectively 250, 200, 50, 50, and 50. Because TerraMind uses a Vision Transformer with a fixed input resolution of 224$\times$224 and patch size of 16, the 500 m window is resized to fit this input, producing pixel dimensions of 14.
The number of samples for training, validation, and testing are respectively 29308, 9253, and 13440.
More information on the dataset collection can be found in \cite{zhou_livability}.

\revdone{We adopt the training, validation, and testing split used by \cite{zhou_livability} rather than constructing our own. Out of the 13 Dutch cities in the dataset, four areas (Eindhoven, Hengelo, Dordrecht, and Beesel) are held out and used for testing. The remaining nine areas (Almere, Amsterdam, Arnhem, Eemsdelta, Groningen, Nijmegen, Rotterdam, Venlo, and Weert) are split across training and validation. 
As such, our reported test performance also measures generalization to entirely unseen cities rather than interpolating within a city. This setting matters as this is a motivating use case for using embeddings in new, unlabeled regions. 
}

\section{Model}
\label{sec:model}

Multimodal deep learning is promising in fusing features from different geospatial modalities~\cite{mmfoundation, mdff, gpmmtransformer}. In our work, we use three primary types of information: images, text, and embeddings. While there has been prior work on multimodal fusion in geospatial tasks~\cite{zhoufuse}, there has been little attention paid to integrating learned embeddings with raster imagery or text. To address this gap, we extend the transformer-based multi-task multimodal regression (TMTMR) model~\cite{zhou_livability} to include additional embedding vectors. 

\begin{figure}[tp]
    \centering
    \includegraphics[width=\linewidth]{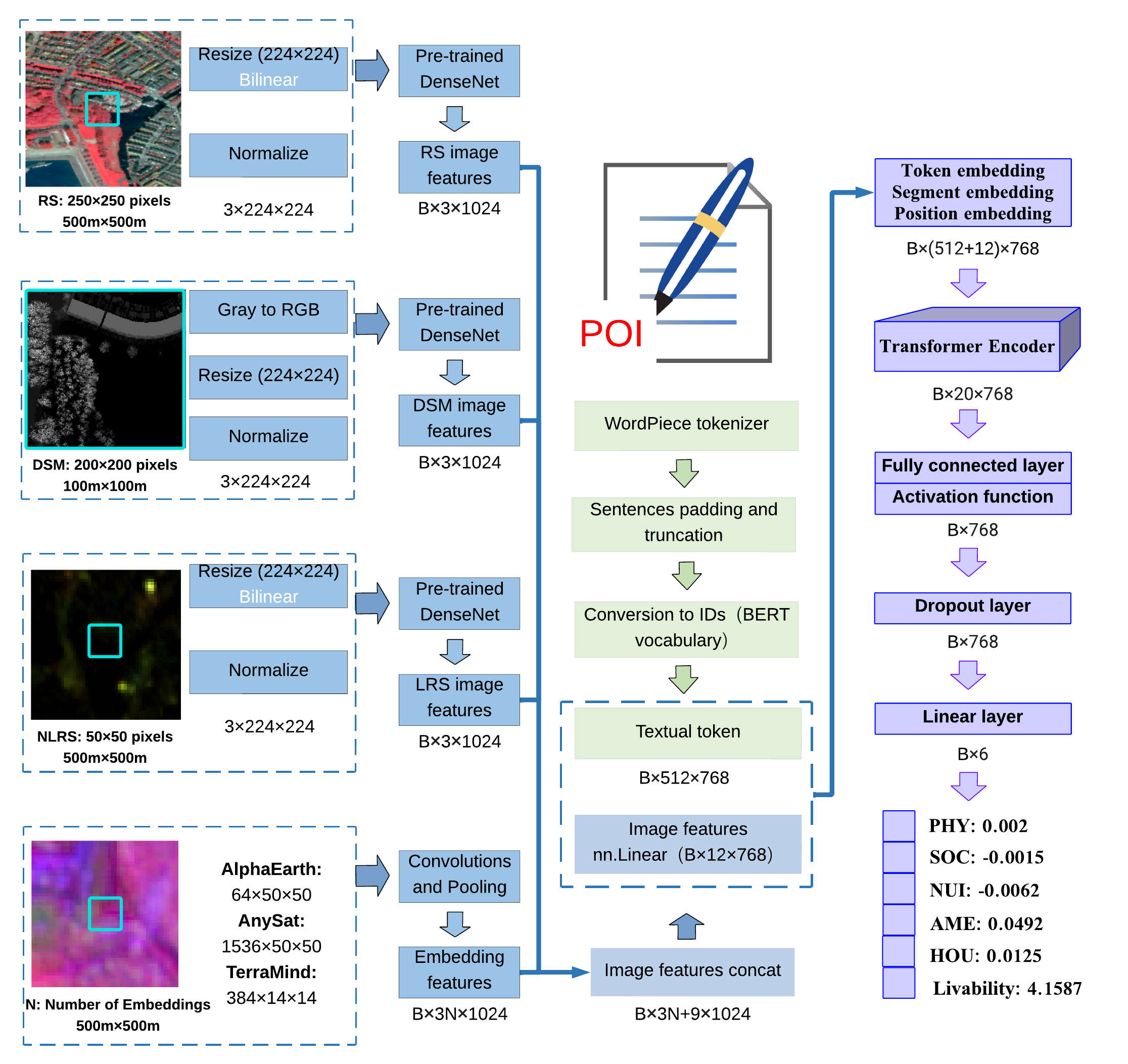}
    \caption{The architecture of the livability study regression model, inspired by \cite{zhou_livability}. Features of RS, DSM, NLRS are extracted by the pretrained DenseNet, features of embeddings are extracted by convolutional layers, and features of POI are extracted by the pretrained BERT network. These features are then concatenated and input into the Transformer Encoder, and finally decoded by a linear layer.}
    \Description{Diagram showing each input being passed through preprocessing steps and then through an encoder. Then, the encoded features are all concatenated and passed through a transformer encoder and then an MP, which then outputs predicted livability scores.}
    \label{fig:model}
\end{figure}

When extending transformers to multimodal inputs or imagery, it is common to extract features from modalities before further processing~\cite{vit, zhoufuse}. Pretrained models are exceptionally important, in this case, due to the lack of training samples. 

For example, pretrained convolutional neural networks (CNNs) are often used as feature extractors for images. Some notable examples include ResNet~\cite{resnet}, EfficientNetV2~\cite{efficientnetv2}, and DenseNet~\cite{densenet}. Of these, DenseNet is significant for its dense block architecture, allowing it to learn both local and global interactions in geospatial applications~\cite{zhoufuse}. As illustrated in \cref{fig:model}, the images are normalized and resized using bilinear interpolation to match the input requirements of DenseNet. The images are then processed through the pretrained DenseNet to extract relevant spatial features.
A powerful feature extractor for text is the Bidirectional Encoder Representation from Transformers (BERT)~\cite{bert}: a pretrained language representation model that has been widely proficient in natural language processing tasks~\cite{bertsurvey}. 
The words in the POI data are first tokenized into subwords and characters using the WordPiece tokenizer. Next, these tokens are truncated or padded to 512 tokens, which is the input size for BERT. Then, each token is mapped to its corresponding ID from the BERT vocabulary. 

Geospatial embeddings, on the other hand, differ from standard images. Each pixel contains a high-dimensional vector rather than three channels. Because there are no suitable pretrained models, we utilize a dedicated series of convolutions and pooling layers to extract essential features from each embedding. We can append additional convolutional feature extractors to add more embeddings as inputs.

All image, embedding, and text features are then merged together before being put into a Transformer Encoder. A decoder comprising linear and dropout layers then outputs the overall livability score and five domain scores. The loss we use is a multitask MAE objective:

\begin{align}
    \mathcal{L} = \frac{1}{N} \sum_{i=1}^N \sum_{t=1}^T \lvert\hat y_{i,t} - y_{i,t}\rvert
\end{align}

where $N$ are the number of samples, $T$ are the number of tasks, $\hat y_{i,t}$ is the predicted score of the $i$-th sample and $y_{i,t}$ is the true score of the $i$-th sample. This allows the model to learn shared features across tasks, harnessing the relationships between modalities. 
\revdeleted{The original paper~\cite{zhou_livability} uses different training parameters, so we do not compare with the original published TMTMR baseline.} More information on the model and training process is available in Appendix~\ref{app:train}.

\begin{table*}[ht]
\centering
\caption{RMSE comparison across livability domain scores between baseline network and models that include foundation model embeddings, \revdone {and linear probes trained on pooled embedding vectors}. Values are mean $\pm$ std over 3 runs. Best values for each score are bolded.}
\label{tab:add}
\revdone{
\begin{tabular}{@{}lcccccc@{}}
\toprule
\textbf{Model} & \textbf{LIV} & \textbf{PHY} & \textbf{NUI} & \textbf{SOC} & \textbf{AME} & \textbf{HOU} \\
\midrule
\textsc{base} & 0.106$\pm$0.002 & 0.027$\pm$0.000 & 0.061$\pm$0.001 & 0.032$\pm$0.001 & 0.055$\pm$0.002 & 0.034$\pm$0.000 \\
\textsc{aef} & \textbf{0.098$\pm$0.001} & \textbf{0.024$\pm$0.000} & \textbf{0.059$\pm$0.000} & \textbf{0.030$\pm$0.000} & 0.047$\pm$0.001 & \textbf{0.033$\pm$0.000} \\
\textsc{as} & 0.104$\pm$0.000 & 0.028$\pm$0.000 & 0.062$\pm$0.001 & 0.032$\pm$0.000 & 0.052$\pm$0.001 & 0.034$\pm$0.000 \\
\textsc{tm} & 0.101$\pm$0.001 & 0.026$\pm$0.000 & 0.060$\pm$0.000 & 0.031$\pm$0.000 & 0.052$\pm$0.001 & 0.034$\pm$0.000 \\
\textsc{aef+as} & 0.100$\pm$0.001 & 0.025$\pm$0.000 & \textbf{0.059$\pm$0.001} & \textbf{0.030$\pm$0.000} & \textbf{0.044$\pm$0.001} & \textbf{0.033$\pm$0.000} \\
\textsc{aef+tm} & \textbf{0.098$\pm$0.000} & 0.025$\pm$0.000 & \textbf{0.059$\pm$0.000} & \textbf{0.030$\pm$0.000} & 0.045$\pm$0.001 & \textbf{0.033$\pm$0.001} \\
\textsc{as+tm} & 0.101$\pm$0.002 & 0.026$\pm$0.000 & 0.060$\pm$0.001 & 0.031$\pm$0.001 & 0.050$\pm$0.001 & 0.034$\pm$0.000 \\
\textsc{aef+as+tm} & \textbf{0.098$\pm$0.001} & \textbf{0.024$\pm$0.000} & \textbf{0.059$\pm$0.000} & \textbf{0.030$\pm$0.000} & \textbf{0.044$\pm$0.001} & \textbf{0.033$\pm$0.001} \\
\midrule
\textsc{probe-aef} & 0.116$\pm$0.002 & \textbf{0.024$\pm$0.001} & 0.064$\pm$0.000 & 0.035$\pm$0.000 & 0.053$\pm$0.001 & 0.037$\pm$0.000 \\
\textsc{probe-as} & 0.130$\pm$0.006 & 0.027$\pm$0.001 & 0.071$\pm$0.002 & 0.040$\pm$0.002 & 0.060$\pm$0.003 & 0.040$\pm$0.002 \\
\textsc{probe-tm} & 0.114$\pm$0.002 & 0.025$\pm$0.001 & 0.064$\pm$0.001 & 0.034$\pm$0.000 & 0.056$\pm$0.002 & 0.035$\pm$0.001 \\
\bottomrule
\end{tabular}
}
\vspace{-1.0em}
\end{table*}

\section{Experiments}
\label{sec:expr}

Using the training dataset described in \Cref{sec:data} and the model illustrated in \Cref{sec:model}, we conducted performance experiments on the effect of including geospatial embeddings. The baseline model (\textsc{base}) includes the original four modalities: RS, DSM, NLRS, and POI data. The augmented models (\textsc{aef}, \textsc{as}, \textsc{tm}) each include the previous four inputs as well as embeddings from AlphaEarth Foundations, AnySat, and TerraMind, respectively. In a few experiments, we include combinations of these embeddings as well to understand whether they provide complementary information or are redundant. Specifically, we evaluate all pairwise combinations (\textsc{aef+as}, \textsc{aef+tm}, \textsc{as+tm}) and the full combination (\textsc{aef+as+tm}) to assess the marginal contribution of each embedding source. 
\revdone{
Following standard foundation model literature~\cite{alphaearth}, we include a baseline embedding-only linear probe for each foundation model (\textsc{probe-aef}, \textsc{probe-as}, \textsc{probe-tm}) to assess how much the multimodal architecture improves the performance. Each probe is trained on the spatially mean-pooled embedding vector and uses the same loss function.
}
We test out different training input combinations and the model's robustness to missing inputs. To compare the performance, we report the RMSE of the six livability scores explained in \Cref{sec:data}.

\revdone{
We compare against the published TMTMR baseline in Appendix~\ref{app:published}, and \textsc{base} trails the published performance by roughly 5 to 11\% RMSE across the six livability scores. We attribute this to training choices like epoch length and loss function rather than modeling decisions. Therefore, all embedding comparisons are relative to our own implementation rather than the published baseline.
}
\revdone{Replication code is available at \url{https://github.com/cybergis/embed-livability}.}


\subsection{Results on adding embeddings}


As illustrated in \Cref{tab:add}, we compare the baseline with augmented models.
\revdeleted{Including AnySat or TerraMind in the model results in similar performances as the baseline model.} \revdone{Including AnySat in the model results in similar performance to the baseline model, while TerraMind provides a small but fairly consistent improvement across scores.} The inclusion of AlphaEarth decreases RMSE in LIV and AME with slight improvements in other scores. This suggests that AlphaEarth's embeddings capture amenity-related information not encoded by AnySat or TerraMind. The AME RMSE further decreases when combining AlphaEarth with other embeddings. This may suggest that AnySat and TerraMind must capture some orthogonal signals that further improve AME performance when combined with AlphaEarth.

\revdone{The embedding-only probes are used to understand how each embedding alone affects the performance. All probes underperform the \textsc{base} and single-embedding models on LIV and NUI, confirming that the multimodal architecture contribute a significant amount of information beyond the embeddings. SOC and HOU are closer with \textsc{probe-tm} having similar RMSE performance to \textsc{base}. 
AME and PHY are interesting domains since \textsc{probe-aef} beats the performance of \textsc{base} in both of these domains.  This indicates that PHY- and AME- relevant information like proximity to infrastructure or amenity density is especially concentrated in these embeddings. }

\revdone{Although combining embeddings improves performance upon linear probes, we caution against over-interpreting this as clean evidence of complementary information. These embeddings are all pretrained on overlapping Earth observation sources, and our geospatial inputs are themselves drawn from related satellite and elevation products. A decrease in RMSE when including multiple embeddings could also be caused by added model capacity or each embedding contributing noisy versions of the same signal, which the model can ensemble into a better estimate. The latter explanation is more likely to be the case due to the performance of the probes. Although \textsc{aef} is clearly stronger than \textsc{tm}, their respective embedding-only probes \textsc{probe-aef} and \textsc{probe-tm} have very similar performance. If AlphaEarth's advantage in the multimodal model simply reflected carrying more raw livability signal, we would expect that to show up in the probe. Instead, the two embeddings are comparable alone but diverge once combined with geospatial inputs. These results are more consistent with AlphaEarth integrating synergistically with the other modalities rather than AlphaEarth have more information.
Disentangling these explanations would require representation-level probing 
or using a formal complementary index~\cite{earthembed_complementarity}
that is beyond the scope of this paper.
}

\begin{table}[ht]
\centering
\caption{RMSE comparison across livability domain scores between baseline network and models that include foundation model embeddings. The city column indicates the city used for testing. Values are mean $\pm$ std over 3 runs. Best values for each score are bolded.
}
\label{tab:city}
\resizebox{1.0\linewidth}{!}{
\revdone{
\begin{tabular}{@{}ll cccccc@{}}
\toprule
\textbf{City} & \textbf{Model} & \textbf{LIV} & \textbf{PHY} & \textbf{NUI} & \textbf{SOC} & \textbf{AME} & \textbf{HOU} \\
\midrule
\multirow{8}{*}{\textsc{eindhoven}} & \textsc{base} & 0.095$\pm$0.002 & 0.025$\pm$0.001 & 0.059$\pm$0.001 & 0.030$\pm$0.001 & 0.054$\pm$0.002 & \textbf{0.028$\pm$0.001} \\
 & \textsc{aef} & 0.089$\pm$0.001 & \textbf{0.022$\pm$0.000} & 0.056$\pm$0.001 & \textbf{0.028$\pm$0.000} & 0.045$\pm$0.001 & \textbf{0.028$\pm$0.000} \\
 & \textsc{as} & 0.096$\pm$0.000 & 0.027$\pm$0.001 & 0.061$\pm$0.001 & 0.030$\pm$0.000 & 0.052$\pm$0.001 & 0.029$\pm$0.000 \\
 & \textsc{tm} & 0.090$\pm$0.002 & 0.024$\pm$0.000 & 0.058$\pm$0.001 & 0.030$\pm$0.000 & 0.050$\pm$0.001 & \textbf{0.028$\pm$0.000} \\
 & \textsc{aef+as} & 0.090$\pm$0.002 & 0.023$\pm$0.000 & 0.056$\pm$0.001 & \textbf{0.028$\pm$0.000} & 0.043$\pm$0.001 & 0.029$\pm$0.001 \\
 & \textsc{aef+tm} & \textbf{0.088$\pm$0.001} & 0.023$\pm$0.001 & \textbf{0.055$\pm$0.001} & \textbf{0.028$\pm$0.000} & 0.044$\pm$0.001 & \textbf{0.028$\pm$0.000} \\
 & \textsc{as+tm} & 0.091$\pm$0.002 & 0.025$\pm$0.000 & 0.059$\pm$0.001 & 0.030$\pm$0.001 & 0.049$\pm$0.002 & \textbf{0.028$\pm$0.001} \\
 & \textsc{aef+as+tm} & 0.090$\pm$0.002 & 0.023$\pm$0.000 & 0.057$\pm$0.001 & \textbf{0.028$\pm$0.001} & \textbf{0.042$\pm$0.001} & \textbf{0.028$\pm$0.001} \\
\midrule
\multirow{8}{*}{\textsc{hengelo}} & \textsc{base} & 0.119$\pm$0.001 & 0.029$\pm$0.000 & 0.058$\pm$0.003 & 0.033$\pm$0.000 & 0.054$\pm$0.002 & 0.040$\pm$0.001 \\
 & \textsc{aef} & \textbf{0.107$\pm$0.001} & 0.026$\pm$0.000 & \textbf{0.056$\pm$0.000} & \textbf{0.030$\pm$0.000} & 0.046$\pm$0.000 & \textbf{0.039$\pm$0.001} \\
 & \textsc{as} & 0.113$\pm$0.001 & 0.028$\pm$0.000 & 0.060$\pm$0.001 & 0.032$\pm$0.000 & 0.050$\pm$0.000 & 0.040$\pm$0.000 \\
 & \textsc{tm} & 0.112$\pm$0.002 & 0.027$\pm$0.000 & 0.059$\pm$0.001 & 0.031$\pm$0.001 & 0.049$\pm$0.001 & 0.040$\pm$0.000 \\
 & \textsc{aef+as} & 0.111$\pm$0.001 & 0.026$\pm$0.001 & 0.059$\pm$0.000 & \textbf{0.030$\pm$0.000} & 0.045$\pm$0.002 & 0.040$\pm$0.001 \\
 & \textsc{aef+tm} & 0.111$\pm$0.001 & 0.026$\pm$0.000 & 0.058$\pm$0.000 & \textbf{0.030$\pm$0.001} & \textbf{0.043$\pm$0.002} & 0.041$\pm$0.001 \\
 & \textsc{as+tm} & 0.111$\pm$0.001 & 0.027$\pm$0.000 & 0.060$\pm$0.002 & 0.032$\pm$0.000 & 0.047$\pm$0.002 & \textbf{0.039$\pm$0.000} \\
 & \textsc{aef+as+tm} & 0.110$\pm$0.001 & \textbf{0.025$\pm$0.000} & 0.058$\pm$0.001 & \textbf{0.030$\pm$0.001} & \textbf{0.043$\pm$0.001} & \textbf{0.039$\pm$0.001} \\
\midrule
\multirow{8}{*}{\textsc{dordrecht}} & \textsc{base} & 0.114$\pm$0.004 & 0.028$\pm$0.000 & 0.068$\pm$0.001 & 0.034$\pm$0.000 & 0.060$\pm$0.002 & 0.039$\pm$0.001 \\
 & \textsc{aef} & 0.106$\pm$0.003 & 0.026$\pm$0.001 & 0.068$\pm$0.001 & 0.034$\pm$0.001 & 0.052$\pm$0.002 & \textbf{0.036$\pm$0.000} \\
 & \textsc{as} & 0.111$\pm$0.001 & 0.028$\pm$0.001 & 0.066$\pm$0.001 & \textbf{0.033$\pm$0.001} & 0.056$\pm$0.001 & 0.039$\pm$0.001 \\
 & \textsc{tm} & 0.110$\pm$0.001 & 0.028$\pm$0.000 & 0.067$\pm$0.001 & 0.034$\pm$0.001 & 0.060$\pm$0.000 & 0.037$\pm$0.000 \\
 & \textsc{aef+as} & 0.109$\pm$0.000 & 0.026$\pm$0.001 & 0.067$\pm$0.001 & \textbf{0.033$\pm$0.000} & \textbf{0.048$\pm$0.001} & 0.037$\pm$0.000 \\
 & \textsc{aef+tm} & \textbf{0.105$\pm$0.002} & 0.026$\pm$0.001 & 0.068$\pm$0.002 & 0.034$\pm$0.000 & 0.051$\pm$0.000 & \textbf{0.036$\pm$0.000} \\
 & \textsc{as+tm} & 0.110$\pm$0.001 & 0.026$\pm$0.000 & \textbf{0.065$\pm$0.002} & \textbf{0.033$\pm$0.001} & 0.055$\pm$0.001 & 0.038$\pm$0.001 \\
 & \textsc{aef+as+tm} & \textbf{0.105$\pm$0.001} & \textbf{0.025$\pm$0.001} & 0.066$\pm$0.001 & \textbf{0.033$\pm$0.000} & 0.049$\pm$0.001 & 0.037$\pm$0.001 \\
\midrule
\multirow{8}{*}{\textsc{beesel}} & \textsc{base} & 0.078$\pm$0.002 & 0.031$\pm$0.002 & \textbf{0.037$\pm$0.003} & 0.028$\pm$0.001 & 0.037$\pm$0.003 & 0.022$\pm$0.001 \\
 & \textsc{aef} & 0.082$\pm$0.006 & \textbf{0.028$\pm$0.001} & 0.041$\pm$0.005 & 0.026$\pm$0.001 & 0.034$\pm$0.005 & 0.021$\pm$0.003 \\
 & \textsc{as} & 0.085$\pm$0.007 & 0.035$\pm$0.001 & 0.043$\pm$0.005 & 0.028$\pm$0.000 & 0.036$\pm$0.002 & 0.024$\pm$0.002 \\
 & \textsc{tm} & 0.080$\pm$0.007 & 0.035$\pm$0.002 & 0.040$\pm$0.005 & 0.027$\pm$0.001 & \textbf{0.032$\pm$0.001} & 0.022$\pm$0.003 \\
 & \textsc{aef+as} & 0.081$\pm$0.009 & \textbf{0.028$\pm$0.001} & 0.039$\pm$0.004 & \textbf{0.024$\pm$0.001} & 0.035$\pm$0.004 & 0.020$\pm$0.001 \\
 & \textsc{aef+tm} & 0.077$\pm$0.002 & \textbf{0.028$\pm$0.001} & 0.039$\pm$0.002 & 0.025$\pm$0.000 & 0.034$\pm$0.003 & \textbf{0.019$\pm$0.001} \\
 & \textsc{as+tm} & 0.080$\pm$0.005 & 0.035$\pm$0.001 & 0.042$\pm$0.002 & 0.026$\pm$0.000 & 0.033$\pm$0.002 & 0.022$\pm$0.001 \\
 & \textsc{aef+as+tm} & \textbf{0.073$\pm$0.002} & 0.029$\pm$0.001 & 0.038$\pm$0.001 & 0.025$\pm$0.001 & 0.036$\pm$0.004 & \textbf{0.019$\pm$0.001} \\
\bottomrule
\end{tabular}
}
}
\vspace{-1.5em}
\end{table}

We split up the testing set into four cities with different traits: Eindhoven (a modern city), Hengelo (a slower-growing city), Dordrecht (a city with a well-preserved historical center), and Beesel (a rural community). The RMSE values for each area are summarized in \Cref{tab:city}. In urban areas (Eindhoven, Hengelo, and Dordrecht), including embeddings generally improves upon the baseline, with AlphaEarth usually providing consistent performance boosts across the livability scores. In Eindhoven, models that include AlphaEarth perform the best, with \textsc{aef} and \textsc{aef+tm} achieving the lowest LIV RMSE scores. In Hengelo and Dordrecht, there is no single embedding combination that consistently outperforms others across all scores. Notably, in Hengelo, the baseline achieves the best NUI RMSE, suggesting that embeddings do not always improve performance on all individual domains.
The rural community of Beesel illustrates a significantly different pattern. Single embeddings, like AlphaEarth and AnySat, worsen LIV performance.
\revdone{Only combinations that pair AlphaEarth and TerraMind, such as \textsc{aef+as+tm}, recover and improve upon it.} 
This implies that embeddings pretrained on predominantly urban geospatial data may not transfer cleanly to rural contexts. Including multiple embeddings may be necessary to mitigate this distributional mismatch.

In \Cref{tab:poi}, we test out how well the baseline and embedding-augmented models perform in areas with and without text information. Consistent with \Cref{tab:add}, \textsc{aef+tm} performs the best across both POI splits. 
All models perform worse with text information, which may be due to increased heterogeneity in dense areas with POI. However, the embeddings still significantly improve performance across the board, especially in areas without POI.
In particular, AME RMSE improves from \revdone{0.048 to 0.040} 
with \textsc{null} POI, compared to \revdone{0.056 to 0.045} 
with \textsc{any} POI. This may reflect AlphaEarth's text-aware pretraining partially substituting for missing POI.

\begin{table}[ht]
\centering
\caption{RMSE comparison across livability domain scores between baseline network and models that include foundation model embeddings. The POI column indicates where there is POI data or not. Values are mean $\pm$ std over 3 runs. Best values for each score are bolded.
}
\label{tab:poi}
\resizebox{1.0\linewidth}{!}{
\revdone{
\begin{tabular}{@{}ll cccccc@{}}
\toprule
\textbf{POI} & \textbf{Model} & \textbf{LIV} & \textbf{PHY} & \textbf{NUI} & \textbf{SOC} & \textbf{AME} & \textbf{HOU} \\
\midrule
\multirow{8}{*}{\textsc{null}} & \textsc{base} & 0.093$\pm$0.001 & 0.031$\pm$0.000 & 0.050$\pm$0.000 & 0.034$\pm$0.000 & 0.048$\pm$0.001 & 0.031$\pm$0.000 \\
 & \textsc{aef} & 0.089$\pm$0.001 & \textbf{0.028$\pm$0.001} & 0.050$\pm$0.001 & 0.032$\pm$0.000 & 0.041$\pm$0.001 & 0.031$\pm$0.000 \\
 & \textsc{as} & 0.095$\pm$0.003 & 0.033$\pm$0.000 & 0.053$\pm$0.002 & 0.034$\pm$0.000 & 0.045$\pm$0.002 & 0.032$\pm$0.000 \\
 & \textsc{tm} & 0.089$\pm$0.002 & 0.031$\pm$0.001 & 0.049$\pm$0.001 & 0.033$\pm$0.000 & 0.045$\pm$0.000 & 0.031$\pm$0.001 \\
 & \textsc{aef+as} & 0.090$\pm$0.002 & 0.029$\pm$0.001 & 0.048$\pm$0.001 & 0.032$\pm$0.000 & 0.041$\pm$0.001 & \textbf{0.030$\pm$0.001} \\
 & \textsc{aef+tm} & \textbf{0.086$\pm$0.000} & 0.029$\pm$0.000 & 0.048$\pm$0.002 & 0.032$\pm$0.000 & \textbf{0.040$\pm$0.001} & 0.031$\pm$0.000 \\
 & \textsc{as+tm} & 0.090$\pm$0.002 & 0.032$\pm$0.001 & 0.050$\pm$0.001 & 0.033$\pm$0.001 & 0.043$\pm$0.002 & 0.032$\pm$0.000 \\
 & \textsc{aef+as+tm} & 0.087$\pm$0.001 & \textbf{0.028$\pm$0.000} & \textbf{0.047$\pm$0.001} & \textbf{0.031$\pm$0.000} & \textbf{0.040$\pm$0.001} & 0.031$\pm$0.000 \\
\midrule
\multirow{8}{*}{\textsc{any}} & \textsc{base} & 0.108$\pm$0.002 & 0.026$\pm$0.001 & 0.062$\pm$0.001 & 0.032$\pm$0.001 & 0.056$\pm$0.002 & 0.035$\pm$0.000 \\
 & \textsc{aef} & \textbf{0.099$\pm$0.001} & 0.024$\pm$0.000 & \textbf{0.060$\pm$0.000} & 0.030$\pm$0.000 & 0.048$\pm$0.001 & \textbf{0.033$\pm$0.000} \\
 & \textsc{as} & 0.105$\pm$0.000 & 0.027$\pm$0.001 & 0.063$\pm$0.001 & 0.031$\pm$0.000 & 0.053$\pm$0.000 & 0.034$\pm$0.000 \\
 & \textsc{tm} & 0.102$\pm$0.001 & 0.025$\pm$0.000 & 0.061$\pm$0.000 & 0.031$\pm$0.000 & 0.053$\pm$0.001 & 0.034$\pm$0.000 \\
 & \textsc{aef+as} & 0.101$\pm$0.001 & 0.024$\pm$0.000 & 0.061$\pm$0.001 & \textbf{0.029$\pm$0.000} & \textbf{0.045$\pm$0.001} & 0.034$\pm$0.000 \\
 & \textsc{aef+tm} & \textbf{0.099$\pm$0.000} & 0.024$\pm$0.001 & \textbf{0.060$\pm$0.000} & 0.030$\pm$0.000 & 0.046$\pm$0.001 & 0.034$\pm$0.001 \\
 & \textsc{as+tm} & 0.102$\pm$0.002 & 0.026$\pm$0.000 & 0.062$\pm$0.001 & 0.031$\pm$0.001 & 0.051$\pm$0.002 & 0.034$\pm$0.000 \\
 & \textsc{aef+as+tm} & 0.100$\pm$0.001 & \textbf{0.023$\pm$0.000} & 0.061$\pm$0.000 & \textbf{0.029$\pm$0.000} & \textbf{0.045$\pm$0.001} & 0.034$\pm$0.001 \\
\bottomrule
\end{tabular}
}
}
\vspace{-1.5em}
\end{table}

\subsection{Results on replacing modalities}

In \Cref{tab:ablation}, we examine the performance when we exclude RS, DSM, NLRS, POI data, or all of the above during training. The goal is to see whether embeddings could mitigate performance degradation when important data sources are missing. Across all single-modality ablations, using any embeddings consistently improves the performance relative to the baseline model.
\revdone{Both \textsc{aef} and \textsc{tm} usually have the largest improvements on \textsc{base}, indicating that AlphaEarth and TerraMind captures a lot of information related to urban livability.}
When the input data excludes DSM, NLRS, and POI, \textsc{aef} recovers similar performance to the full-data models trained in \Cref{tab:add}. The RS ablation is slightly more difficult: \textsc{aef} does improve upon \textsc{base} in this setting, but it still has worse LIV RMSE than the full-data \textsc{aef} network. This reflects the fine-grained information not captured by embeddings but reflected in RS imagery.


We further examine if the model can solely learn from combinations of embedding without RS, DSM, NLRS, and POI data. Out of these, the highest performing model is \textsc{aef+tm}, and it achieves similar scores to the full-data \textsc{base} model on all metrics and even outperforms some domains like PHY and AME.
Incorporating AnySat into \textsc{aef+tm} hurts RMSE, so AnySat may introduce redundant or noisy information without other modalities.


\begin{table}[ht]
\centering
\caption{RMSE comparison across livability domain scores between baseline network and models that include foundation model embeddings. The ablation column indicates the modality excluded during training. Values are mean $\pm$ std over 3 runs. Best values for each score are bolded.
}
\label{tab:ablation}
\resizebox{1.0\linewidth}{!}{
\revdone{
\begin{tabular}{@{}ll cccccc@{}}
\toprule
\textbf{Ablation} & \textbf{Model} & \textbf{LIV} & \textbf{PHY} & \textbf{NUI} & \textbf{SOC} & \textbf{AME} & \textbf{HOU} \\
\midrule
\multirow{4}{*}{RS} & \textsc{base} & 0.122$\pm$0.000 & 0.029$\pm$0.000 & 0.067$\pm$0.001 & 0.036$\pm$0.000 & 0.059$\pm$0.002 & 0.038$\pm$0.000 \\
 & \textsc{aef} & \textbf{0.105$\pm$0.001} & \textbf{0.024$\pm$0.000} & \textbf{0.062$\pm$0.000} & \textbf{0.032$\pm$0.000} & \textbf{0.047$\pm$0.002} & \textbf{0.035$\pm$0.000} \\
 & \textsc{as} & 0.110$\pm$0.001 & 0.028$\pm$0.001 & 0.065$\pm$0.001 & 0.034$\pm$0.000 & 0.055$\pm$0.001 & 0.037$\pm$0.000 \\
 & \textsc{tm} & \textbf{0.105$\pm$0.002} & 0.026$\pm$0.000 & \textbf{0.062$\pm$0.001} & \textbf{0.032$\pm$0.000} & 0.054$\pm$0.000 & \textbf{0.035$\pm$0.000} \\
\midrule
\multirow{4}{*}{DSM} & \textsc{base} & 0.104$\pm$0.002 & 0.028$\pm$0.001 & 0.062$\pm$0.000 & 0.032$\pm$0.000 & 0.057$\pm$0.000 & 0.034$\pm$0.000 \\
 & \textsc{aef} & \textbf{0.100$\pm$0.002} & \textbf{0.024$\pm$0.000} & \textbf{0.060$\pm$0.001} & \textbf{0.031$\pm$0.000} & \textbf{0.048$\pm$0.001} & \textbf{0.033$\pm$0.001} \\
 & \textsc{as} & 0.103$\pm$0.000 & 0.028$\pm$0.001 & 0.062$\pm$0.000 & \textbf{0.031$\pm$0.001} & 0.054$\pm$0.001 & \textbf{0.033$\pm$0.000} \\
 & \textsc{tm} & \textbf{0.100$\pm$0.001} & 0.027$\pm$0.000 & \textbf{0.060$\pm$0.001} & \textbf{0.031$\pm$0.000} & 0.052$\pm$0.000 & \textbf{0.033$\pm$0.000} \\
\midrule
\multirow{4}{*}{NLRS} & \textsc{base} & 0.108$\pm$0.001 & 0.029$\pm$0.000 & 0.063$\pm$0.001 & 0.033$\pm$0.000 & 0.059$\pm$0.001 & \textbf{0.034$\pm$0.001} \\
 & \textsc{aef} & \textbf{0.101$\pm$0.003} & \textbf{0.025$\pm$0.000} & \textbf{0.061$\pm$0.001} & \textbf{0.031$\pm$0.001} & \textbf{0.049$\pm$0.002} & \textbf{0.034$\pm$0.001} \\
 & \textsc{as} & 0.107$\pm$0.003 & 0.028$\pm$0.001 & 0.063$\pm$0.001 & 0.032$\pm$0.000 & 0.057$\pm$0.002 & 0.035$\pm$0.000 \\
 & \textsc{tm} & 0.102$\pm$0.002 & 0.027$\pm$0.000 & \textbf{0.061$\pm$0.002} & \textbf{0.031$\pm$0.000} & 0.053$\pm$0.001 & \textbf{0.034$\pm$0.000} \\
\midrule
\multirow{4}{*}{POI} & \textsc{base} & 0.111$\pm$0.017 & 0.027$\pm$0.001 & 0.067$\pm$0.010 & 0.037$\pm$0.010 & 0.062$\pm$0.003 & 0.037$\pm$0.005 \\
 & \textsc{aef} & 0.099$\pm$0.002 & \textbf{0.025$\pm$0.000} & 0.060$\pm$0.001 & \textbf{0.030$\pm$0.000} & \textbf{0.048$\pm$0.002} & 0.034$\pm$0.000 \\
 & \textsc{as} & 0.103$\pm$0.002 & 0.028$\pm$0.000 & 0.062$\pm$0.001 & 0.032$\pm$0.000 & 0.054$\pm$0.002 & 0.034$\pm$0.000 \\
 & \textsc{tm} & \textbf{0.098$\pm$0.001} & 0.026$\pm$0.000 & \textbf{0.059$\pm$0.000} & \textbf{0.030$\pm$0.001} & 0.056$\pm$0.001 & \textbf{0.033$\pm$0.000} \\
\midrule
\multirow{7}{*}{\makecell[l]{RS\\DSM\\NLRS\\POI}} & \textsc{aef} & 0.112$\pm$0.001 & \textbf{0.024$\pm$0.001} & 0.065$\pm$0.001 & 0.034$\pm$0.000 & 0.051$\pm$0.002 & 0.036$\pm$0.001 \\
 & \textsc{as} & 0.115$\pm$0.002 & 0.027$\pm$0.001 & 0.067$\pm$0.001 & 0.035$\pm$0.000 & 0.064$\pm$0.001 & 0.037$\pm$0.000 \\
 & \textsc{tm} & 0.109$\pm$0.001 & 0.025$\pm$0.000 & 0.063$\pm$0.001 & 0.033$\pm$0.000 & 0.056$\pm$0.001 & \textbf{0.035$\pm$0.000} \\
 & \textsc{aef+as} & 0.113$\pm$0.001 & \textbf{0.024$\pm$0.000} & 0.065$\pm$0.001 & 0.033$\pm$0.000 & \textbf{0.050$\pm$0.003} & 0.036$\pm$0.000 \\
 & \textsc{aef+tm} & \textbf{0.107$\pm$0.001} & \textbf{0.024$\pm$0.000} & \textbf{0.062$\pm$0.001} & \textbf{0.032$\pm$0.000} & \textbf{0.050$\pm$0.002} & 0.036$\pm$0.000 \\
 & \textsc{as+tm} & 0.109$\pm$0.002 & 0.026$\pm$0.000 & 0.066$\pm$0.001 & 0.034$\pm$0.000 & 0.059$\pm$0.003 & 0.036$\pm$0.000 \\
 & \textsc{aef+as+tm} & 0.110$\pm$0.003 & 0.026$\pm$0.002 & 0.063$\pm$0.001 & 0.033$\pm$0.003 & \textbf{0.050$\pm$0.005} & 0.036$\pm$0.002 \\
\bottomrule
\end{tabular}
}
}
\vspace{-1.5em}
\end{table}

\subsection{Results on missing modality}

We also examine how the trained full-data models in \Cref{tab:add} respond when one of their inputs is missing to understand how models weigh different modalities. In \Cref{tab:zeroing_out}, we set one of the modalities to zero and calculate the resulting RMSE. Models that include AlphaEarth tend to be more robust towards this ablation, so these embeddings would most effectively compensate for missing modalities.
\revdone{In particular, \textsc{aef+as+tm} tends to be robust with missing modalities as it likely places more weight on the foundation model embeddings.} 
The largest performance gap between the baseline and augmented models occur when RS is missing. This indicates that RS carries the most unique information, which is consistent with related studies~\cite{zhou_livability}.
Notably, zeroing out POI yields a counterintuitive result: \textsc{base} achieves a LIV RMSE of \revdone{0.099, lower than its full-data score of 0.106 (\Cref{tab:add})}. 
This is consistent with the POI stratification results in \Cref{tab:poi} (single-seed), where \textsc{base} models perform worse in areas with POI present (0.105) compared to areas without POI (0.092). Together, these results suggest that POI data introduces heterogeneity that complicates overall livability prediction, even while improving domain-specific scores like AME, likely because dense POI areas are more spatially variable and harder to predict uniformly.

\begin{table}[ht]
\centering
\caption{RMSE comparison across livability domain scores between baseline network and models that include foundation model embeddings. All models are trained on RS, DSM, NLRS, POI, and the specified embeddings. The ablation column indicates what modality was zeroed out during testing. Values are mean $\pm$ std over 3 runs. Best values for each score are bolded.}
\label{tab:zeroing_out}
\resizebox{1.0\linewidth}{!}{
\revdone{
\begin{tabular}{@{}ll cccccc@{}}
\toprule
\textbf{Ablation} & \textbf{Model} & \textbf{LIV} & \textbf{PHY} & \textbf{NUI} & \textbf{SOC} & \textbf{AME} & \textbf{HOU} \\
\midrule
\multirow{8}{*}{RS} & \textsc{base} & 0.156$\pm$0.007 & 0.039$\pm$0.004 & 0.076$\pm$0.002 & 0.049$\pm$0.003 & 0.068$\pm$0.004 & 0.050$\pm$0.007 \\
 & \textsc{aef} & 0.112$\pm$0.003 & 0.026$\pm$0.000 & 0.063$\pm$0.002 & 0.034$\pm$0.001 & 0.048$\pm$0.001 & 0.036$\pm$0.001 \\
 & \textsc{as} & 0.138$\pm$0.006 & 0.038$\pm$0.003 & 0.074$\pm$0.003 & 0.044$\pm$0.003 & 0.066$\pm$0.002 & 0.041$\pm$0.001 \\
 & \textsc{tm} & 0.117$\pm$0.002 & 0.030$\pm$0.001 & 0.066$\pm$0.001 & 0.039$\pm$0.002 & 0.063$\pm$0.001 & 0.038$\pm$0.001 \\
 & \textsc{aef+as} & 0.111$\pm$0.002 & 0.026$\pm$0.001 & 0.062$\pm$0.001 & 0.033$\pm$0.001 & 0.045$\pm$0.001 & 0.036$\pm$0.001 \\
 & \textsc{aef+tm} & \textbf{0.108$\pm$0.001} & 0.026$\pm$0.001 & \textbf{0.061$\pm$0.001} & 0.033$\pm$0.000 & 0.046$\pm$0.001 & 0.036$\pm$0.001 \\
 & \textsc{as+tm} & 0.119$\pm$0.010 & 0.031$\pm$0.003 & 0.069$\pm$0.005 & 0.040$\pm$0.005 & 0.063$\pm$0.007 & 0.039$\pm$0.003 \\
 & \textsc{aef+as+tm} & \textbf{0.108$\pm$0.001} & \textbf{0.025$\pm$0.001} & 0.062$\pm$0.001 & \textbf{0.032$\pm$0.001} & \textbf{0.044$\pm$0.001} & \textbf{0.035$\pm$0.000} \\
\midrule
\multirow{8}{*}{DSM} & \textsc{base} & 0.107$\pm$0.002 & 0.033$\pm$0.002 & 0.061$\pm$0.001 & 0.032$\pm$0.001 & 0.059$\pm$0.001 & 0.034$\pm$0.001 \\
 & \textsc{aef} & \textbf{0.098$\pm$0.001} & 0.026$\pm$0.001 & \textbf{0.059$\pm$0.000} & \textbf{0.030$\pm$0.000} & 0.047$\pm$0.001 & \textbf{0.033$\pm$0.000} \\
 & \textsc{as} & 0.103$\pm$0.001 & 0.030$\pm$0.001 & 0.062$\pm$0.002 & 0.032$\pm$0.000 & 0.056$\pm$0.001 & \textbf{0.033$\pm$0.000} \\
 & \textsc{tm} & 0.101$\pm$0.001 & 0.029$\pm$0.000 & 0.060$\pm$0.000 & 0.031$\pm$0.000 & 0.055$\pm$0.001 & 0.034$\pm$0.000 \\
 & \textsc{aef+as} & 0.100$\pm$0.001 & 0.026$\pm$0.001 & 0.060$\pm$0.001 & \textbf{0.030$\pm$0.000} & 0.045$\pm$0.000 & \textbf{0.033$\pm$0.000} \\
 & \textsc{aef+tm} & \textbf{0.098$\pm$0.001} & 0.026$\pm$0.001 & \textbf{0.059$\pm$0.001} & \textbf{0.030$\pm$0.000} & 0.046$\pm$0.001 & \textbf{0.033$\pm$0.000} \\
 & \textsc{as+tm} & 0.101$\pm$0.002 & 0.029$\pm$0.001 & 0.061$\pm$0.002 & 0.031$\pm$0.000 & 0.054$\pm$0.001 & \textbf{0.033$\pm$0.001} \\
 & \textsc{aef+as+tm} & 0.099$\pm$0.001 & \textbf{0.025$\pm$0.000} & 0.060$\pm$0.000 & \textbf{0.030$\pm$0.000} & \textbf{0.044$\pm$0.001} & \textbf{0.033$\pm$0.001} \\
\midrule
\multirow{8}{*}{NLRS} & \textsc{base} & 0.111$\pm$0.002 & 0.039$\pm$0.001 & 0.064$\pm$0.001 & \textbf{0.033$\pm$0.000} & 0.070$\pm$0.002 & 0.035$\pm$0.000 \\
 & \textsc{aef} & \textbf{0.104$\pm$0.002} & 0.027$\pm$0.001 & \textbf{0.061$\pm$0.000} & \textbf{0.033$\pm$0.001} & 0.048$\pm$0.001 & \textbf{0.034$\pm$0.000} \\
 & \textsc{as} & 0.107$\pm$0.002 & 0.038$\pm$0.002 & 0.064$\pm$0.002 & \textbf{0.033$\pm$0.000} & 0.063$\pm$0.003 & 0.035$\pm$0.000 \\
 & \textsc{tm} & 0.109$\pm$0.002 & 0.035$\pm$0.000 & 0.062$\pm$0.001 & \textbf{0.033$\pm$0.000} & 0.063$\pm$0.004 & 0.035$\pm$0.000 \\
 & \textsc{aef+as} & 0.105$\pm$0.001 & 0.027$\pm$0.001 & 0.062$\pm$0.001 & \textbf{0.033$\pm$0.001} & 0.046$\pm$0.000 & 0.035$\pm$0.000 \\
 & \textsc{aef+tm} & \textbf{0.104$\pm$0.002} & 0.028$\pm$0.001 & 0.062$\pm$0.001 & \textbf{0.033$\pm$0.001} & 0.046$\pm$0.001 & \textbf{0.034$\pm$0.001} \\
 & \textsc{as+tm} & 0.107$\pm$0.001 & 0.035$\pm$0.002 & 0.062$\pm$0.001 & \textbf{0.033$\pm$0.001} & 0.059$\pm$0.004 & 0.035$\pm$0.000 \\
 & \textsc{aef+as+tm} & \textbf{0.104$\pm$0.001} & \textbf{0.026$\pm$0.002} & 0.062$\pm$0.001 & \textbf{0.033$\pm$0.000} & \textbf{0.045$\pm$0.001} & \textbf{0.034$\pm$0.000} \\
\midrule
\multirow{8}{*}{POI} & \textsc{base} & 0.099$\pm$0.002 & 0.027$\pm$0.000 & 0.063$\pm$0.001 & 0.033$\pm$0.000 & 0.069$\pm$0.003 & 0.034$\pm$0.000 \\
 & \textsc{aef} & 0.098$\pm$0.001 & \textbf{0.025$\pm$0.001} & 0.061$\pm$0.000 & 0.031$\pm$0.001 & 0.053$\pm$0.001 & 0.034$\pm$0.000 \\
 & \textsc{as} & 0.099$\pm$0.001 & 0.027$\pm$0.000 & 0.063$\pm$0.001 & 0.032$\pm$0.000 & 0.065$\pm$0.003 & 0.034$\pm$0.000 \\
 & \textsc{tm} & 0.097$\pm$0.002 & 0.026$\pm$0.001 & 0.062$\pm$0.002 & 0.032$\pm$0.000 & 0.058$\pm$0.001 & 0.034$\pm$0.000 \\
 & \textsc{aef+as} & 0.099$\pm$0.001 & \textbf{0.025$\pm$0.000} & 0.061$\pm$0.002 & 0.030$\pm$0.000 & 0.049$\pm$0.001 & \textbf{0.033$\pm$0.000} \\
 & \textsc{aef+tm} & \textbf{0.096$\pm$0.002} & 0.026$\pm$0.001 & \textbf{0.059$\pm$0.001} & 0.030$\pm$0.001 & 0.050$\pm$0.003 & 0.034$\pm$0.001 \\
 & \textsc{as+tm} & \textbf{0.096$\pm$0.000} & 0.027$\pm$0.001 & 0.062$\pm$0.001 & 0.032$\pm$0.000 & 0.055$\pm$0.004 & 0.034$\pm$0.000 \\
 & \textsc{aef+as+tm} & 0.098$\pm$0.001 & \textbf{0.025$\pm$0.000} & 0.061$\pm$0.001 & \textbf{0.029$\pm$0.001} & \textbf{0.048$\pm$0.002} & 0.034$\pm$0.000 \\
\bottomrule
\end{tabular}
}
}
\vspace{-1.5em}
\end{table}

\subsection{Geographic Variation}

\begin{figure*}[tp]
    \centering
    \includegraphics[width=\textwidth]{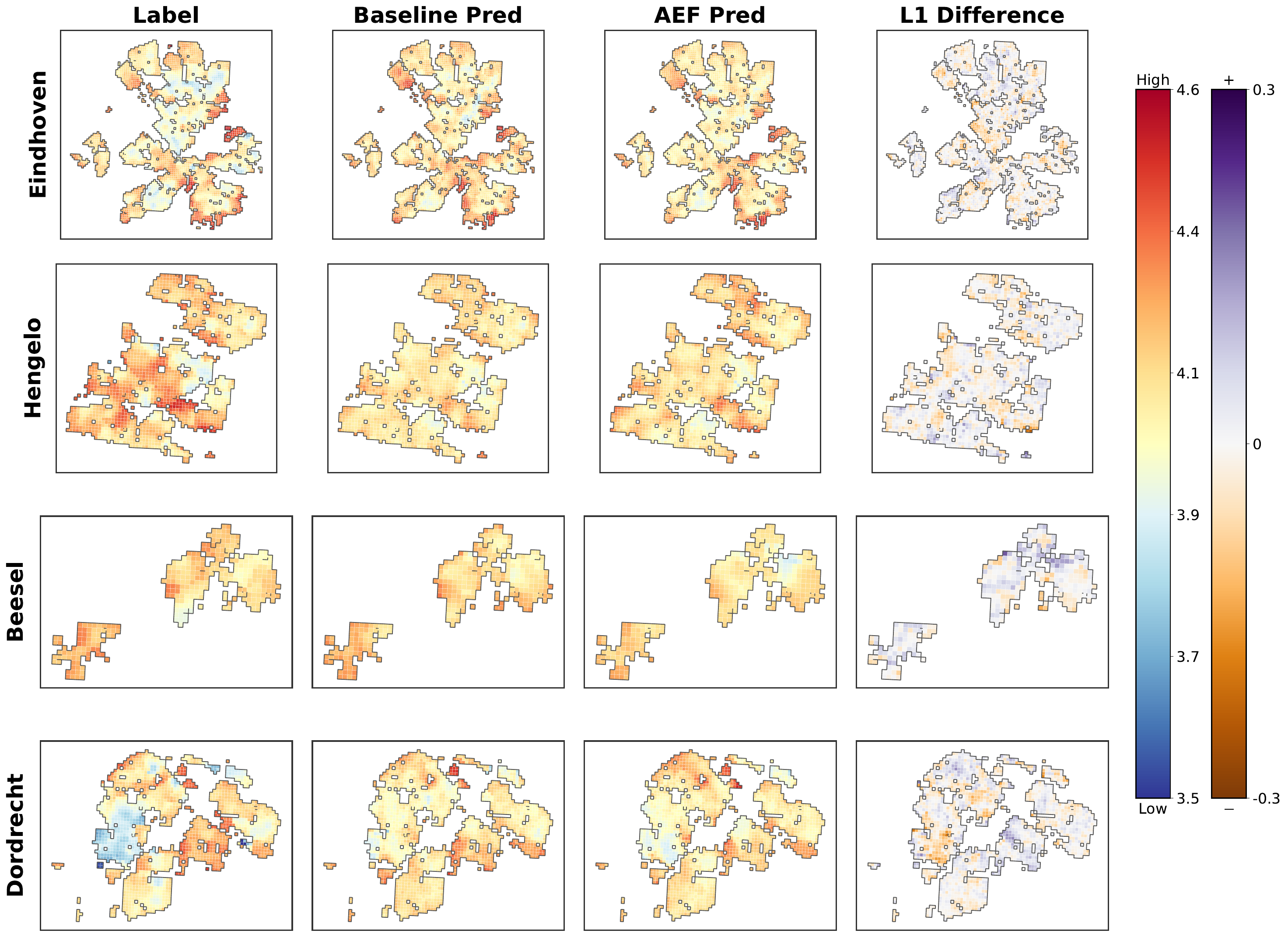}
    \caption{The overall livability score map of several test cities: Eindhoven, Hengelo, Beesel, and Dordrecht. From left to right, we present: the ground truth label, the \textsc{base} prediction, the \textsc{aef} prediction, and the difference between the L1 scores of the \textsc{base} and \textsc{aef} models. The left three columns correspond to the leftmost colorbar. The rightmost column corresponds to the rightmost colorbar, where higher scores indicate \textsc{aef} has a higher L1 than \textsc{base}.
    }
    \Description{A 4 $\times$ 4 plot of different test cities. The rows correspond to Eindhoven, Hegnelo, Beesel, and Dordrecht from top to bottom. The columns correspond to the LIV score, prediction from \textsc{base}, prediction from \textsc{aef}, and the the difference between the L1 scores of the \textsc{base} and \textsc{aef} models. All plots are a 2D heatmap plotted on each city.}
    \label{fig:cities}
\end{figure*}

Illustrated in \Cref{fig:cities}, we examine how the full-data \textsc{base} and \textsc{aef} predictions differ spatially across four test cities. In the rightmost column, we plot the difference between the LIV L1 scores for both \textsc{base} and \textsc{aef}. Orange indicates regions where \textsc{aef} has lower L1 than \textsc{base}, and purple indicates where \textsc{base} performs better. In Eindhoven and Hengelo, the difference maps show small but consistent orange regions, which illustrates that AlphaEarth improves predictions across most urban areas. Dordrecht shows a mixed pattern, with purple concentrated near the historical center. In contrast, Beesel is predominantly purple, confirming that \textsc{aef} underperforms \textsc{base} in this rural community, consistent with the quantitative results in \Cref{tab:city}.

\section{Discussion}
\label{sec:discussion}

This following section consists of explainability techniques to understand how the model weighs AlphaEarth versus other modalities. As such, we analyze the Transformer attention mechanism and its entropy structure, as well as gradient-based activation maps within the convolutional branches.

\subsection{Attention}

\begin{figure*}[t]
    \centering
    \includegraphics[width=0.45\textwidth]{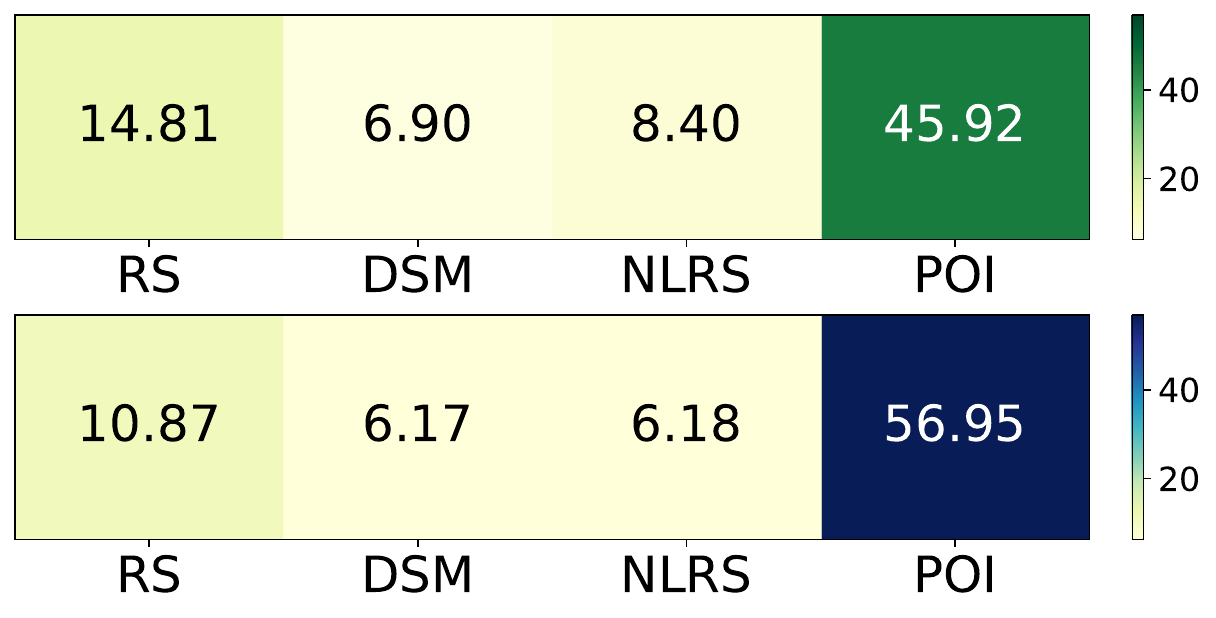}
    \hfill
    \includegraphics[width=0.45\textwidth]{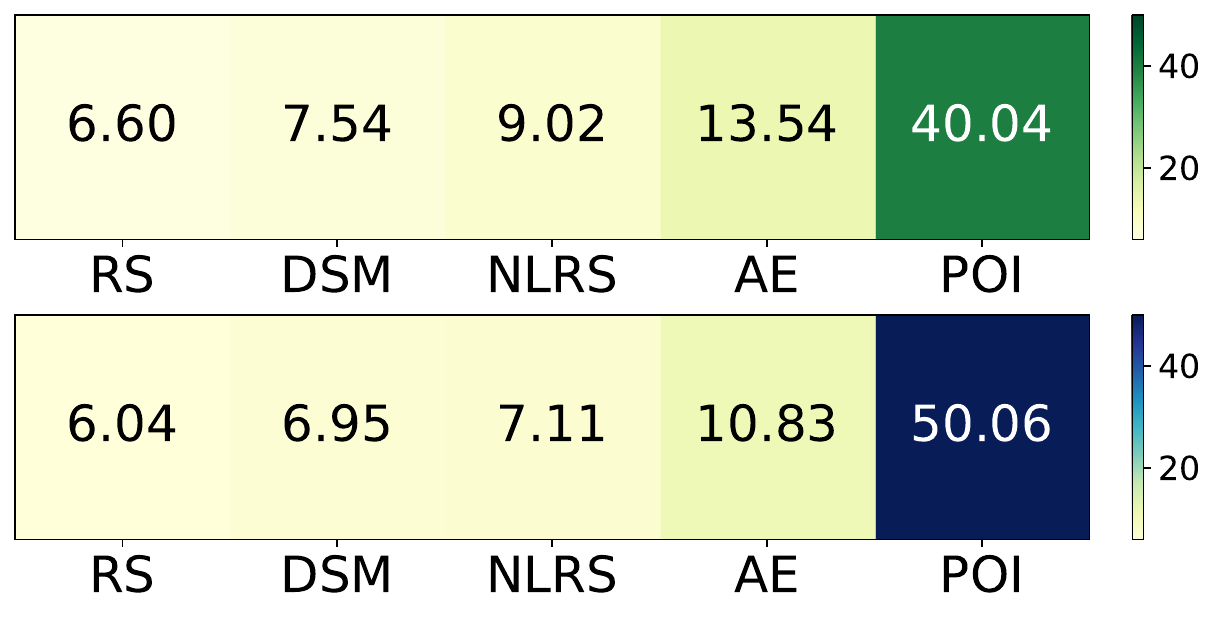}
    \caption{Average attention weights for the top and bottom 50\% of samples by true overall livability scores. The top and bottom rows correspond to the high and low livability groups, respectively. The left and right heatmaps correspondingly refer to the \textsc{base} and \textsc{aef} models. AE corresponds to AlphaEarth embeddings.}
    \Description{This is a 2 $\times$ 2 plot of the average attention weights for high and low livability samples. We report these for a \textsc{base} and \textsc{aef} model. For the \textsc{base} model and high livability samples, the RS, DSM, NLRS, and POI average attention weights are 14.81, 6.90, 8.40, 45.92. For the low livability samples, it becomes 10.87, 6.17, 6.18, and 56.95. For the \textsc{aef} model and high livability areas, the RS, DSM, NLRS, AEF, and POI average attention weights are 6.60, 7.54, 9.02, 13.54, and 40.04 respectively. For low livability samples, the average attention weights are 6.04, 6.95, 7.11, 10.83, and 50.06.}
    \label{fig:attention}
\end{figure*}

Transformers use attention to weigh the relative importance of different modalities. We can use this to gain insight into how the model processes AlphaEarth embeddings. Notably, high attention can have two interpretations: either the modality is crucial, or there is significant heterogeneity within the modality. In the latter case, the model needs to learn the internal differences within that modality, causing the transformer to assign high attention. With this in mind, \Cref{fig:attention} displays average attention heatmaps for \textsc{base} and \textsc{aef} models in 2 distinct groups: high and low livability areas. 

In the figure, the RS and POI attention scores both decrease when AlphaEarth is included, while DSM and NLRS attention scores increase. One interpretation is that AlphaEarth encodes information that overlaps with RS and POI, leading the model to redistribute attention away from those modalities. However, attention weights alone do not establish this as high attention may reflect internal heterogeneity.
Consistent with this caution, the missing-modality results in \Cref{tab:zeroing_out} show that \textsc{aef} substantially reduces the RMSE penalty when RS is zeroed out (from \revdone{0.156 to 0.112} 
for LIV), lending indirect support to the view that AlphaEarth partially compensates for missing RS information.

\revdone{Both \textsc{base} and \textsc{aef} assign POI to have the highest attention of any modality in \Cref{fig:attention}, even though POI gives less information for livability prediction. This is backed up since the performance drop when we zero-out POI isn't as significant as RS or NLRS in \Cref{tab:zeroing_out}. Consistent with the explanation proposed for the same phenomenon in \cite{zhou_livability}, POI draws the most attention because the model must learn to resolve substantial internal variation in the text from empty strings to dense, heterogeneous facility listings. In addition, POI is the modality most directly related to AME, so part of POI's attention may reflect the model being important for one of the domains rather than broadly important across all outputs.
}

To understand how the attention is used differently between the \textsc{base} and \textsc{aef} models, we compute the entropy of the
attention distribution across modalities for each sample. High-livability areas
exhibit significantly higher attention entropy in both the \textsc{base} and
\textsc{aef} models, indicating that attention is more broadly distributed across modalities rather
than concentrated.
\revdone{This entropy gap is small in absolute terms but consistent across both models (\textsc{base} $+0.012$, 95\% CI $[0.008, 0.016]$, $p = 1.65\times10^{-10}$; \textsc{aef} $+0.022$, 95\% CI $[0.019, 0.026]$, $p = 1.71\times10^{-32}$).}
Furthermore, \textsc{aef} attention weights are positively
correlated with both LIV scores (Pearson $r = 0.25$\revdone{, 95\% CI $[0.23, 0.26]$, $p = 2.27\times10^{-186}$}) and attention entropy
($r = 0.26$\revdone{, 95\% CI $[0.25, 0.28]$, $p = 1.69\times10^{-209}$}), suggesting that high-livability areas exhibit greater input
heterogeneity across spatial modalities, prompting the model to draw on a 
broader set of embeddings rather than relying more heavily on \textsc{aef} 
specifically. In low-livability areas, POI dominates attention in both models, 
whereas high-livability areas distribute attention more evenly across RS, NLRS, 
and \textsc{aef}.




\subsection{Full Grad-CAM}
\label{sec:gradcam}

We apply Full Grad-CAM to understand how the inclusion of AlphaEarth changes the spatial attention of the model across the RS, DSM, and NLRS branches. Gradient-weighted Class Activation Mapping (Grad-CAM)~\cite{gradcam} is an explainability technique
to describe the decision-making process of convolutional neural networks. It involves using the gradient of the specified target concept to produce a coarse localization map, highlighting the most significant areas in the image. Standard Grad-CAM applies a ReLU to the weighted activation maps, only allowing positive contributions to the target output. We instead use Full Grad-CAM, which removes this ReLU operation, allowing both positive and negative feature contributions to be visualized. The core form for Full Grad-CAM is:

\begin{equation}
    L(x, y) = \sum_k \alpha_k^c A^k(x,y) 
\end{equation}

where $A^k(x,y)$ is the activation value of the $k$-th convolutional feature map at spatial location $(x,y)$, and $\alpha_k^c$ represents the gradient-based weight for the $k$-th feature map with respect to the $c$-th regression output. 

\begin{figure*}[t]
    \centering
    \includegraphics[width=0.45\textwidth]{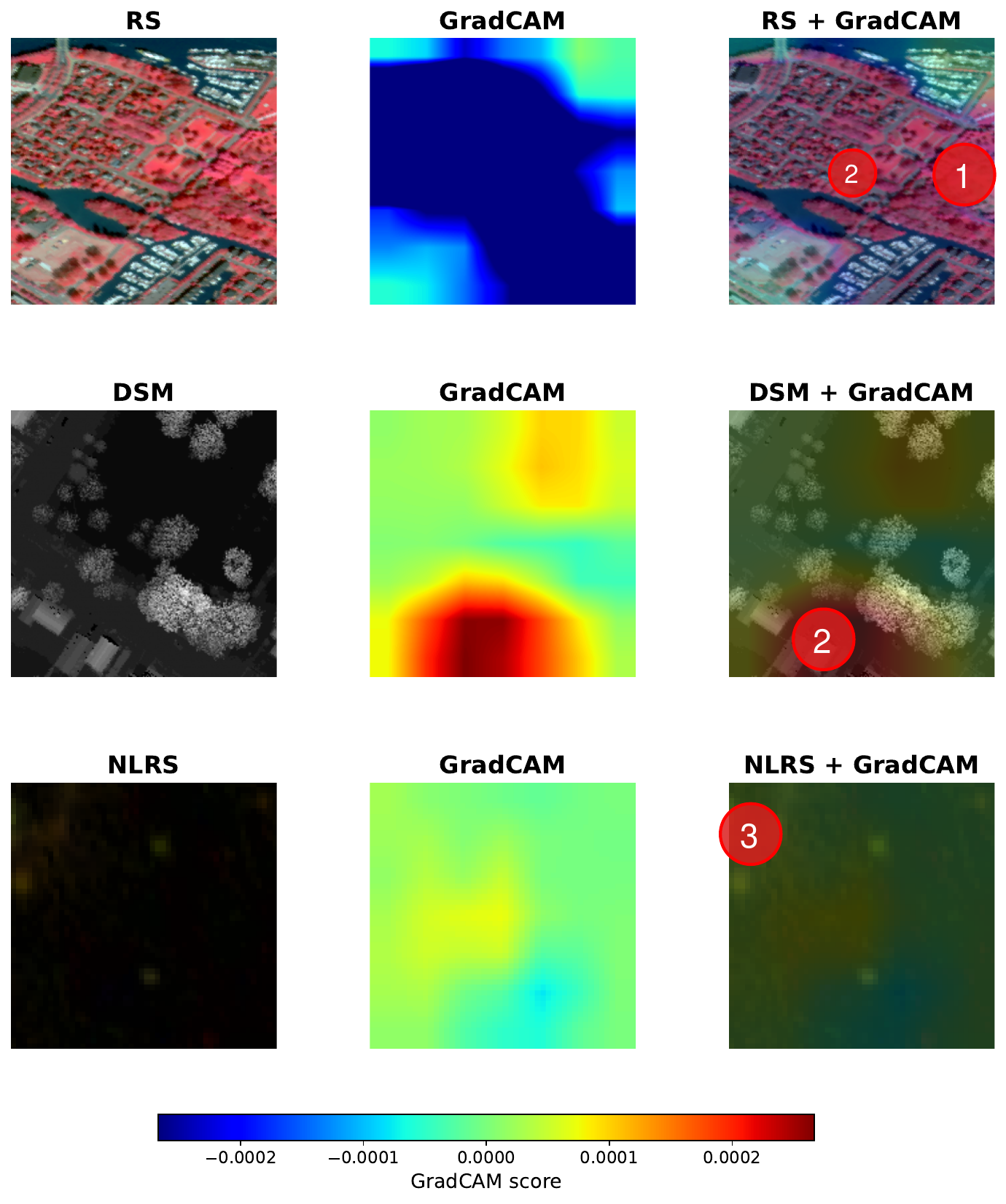}
    \hfill
    \includegraphics[width=0.45\textwidth]{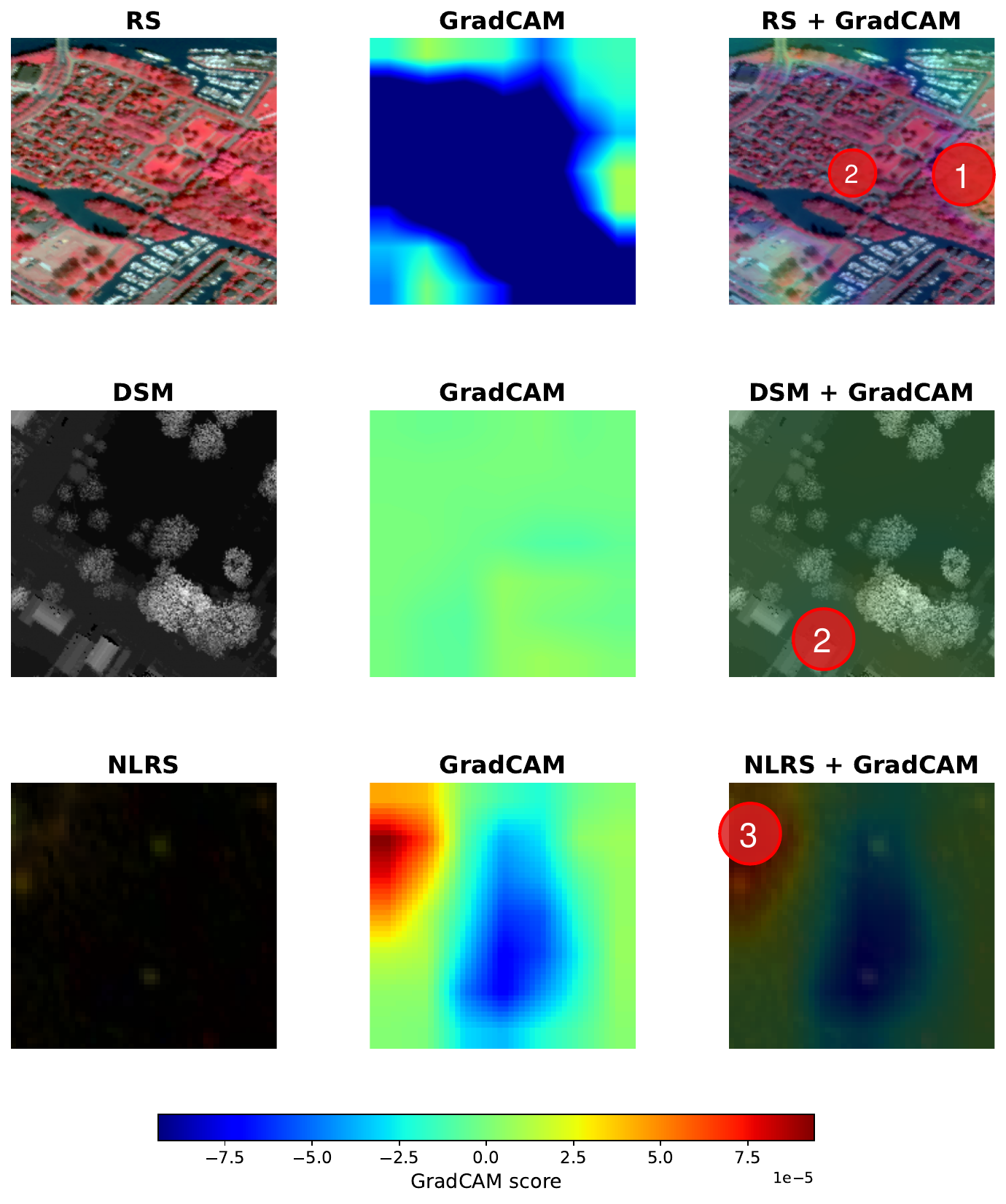}
    \caption{A sample for full Grad-CAM visualization for overall livability scores. The left and right Grad-CAM scores respectively correspond to the \textsc{base} and \textsc{aef} models. The red circles correspond to important areas that affect livability, and they are explained in Section~\ref{sec:gradcam}.}
    \Description{This shows a heatmap for a specific sample in the Netherlands. The left includes the raw RS, DSM, and NLRS image, the \textsc{base} GradCAM heatmap, and the overlap of the two. The right includes the same raw images, the \textsc{aef} GradCAM heatmap, and the overlap of the two. There are numbers one, two, and three on specific parts of the images, which are referenced in \Cref{sec:gradcam}.}
    \label{fig:gradcam}
\end{figure*}

In \Cref{fig:gradcam}, we examine how Full Grad-CAM differs between the \textsc{base} and \textsc{aef} models on a test sample in the city of Dordrecht. The inclusion of AlphaEarth produces notable changes in how the model attends to specific regions across all three modalities, which we highlight with three annotated examples. In particular, the section highlighted \redcircle{1} is a park. The \textsc{aef} model has a much stronger relative Grad-CAM score in this zone, suggesting that the inclusion of global embeddings enables the model to identify structures near the image edge. In DSM, the \redcircle{2} region, corresponding to a residential home, has a much stronger positive Grad-CAM response in the \textsc{base} model. This is inconsistent with the RS Grad-CAM where both models already assigned the same region with a negative contribution. The \textsc{aef} model resolves this inconsistency, suppressing the positive DSM response and aligning it with the RS signal, suggesting that AlphaEarth helps the model integrate information more consistently across modalities. The final area denoted by \redcircle{3} is an example of how the \textsc{aef} model is able to attend to regions near the border in NLRS. This particular section corresponds to an old water tower that has been turned into a landmark with surrounding walkable gardens. AlphaEarth's pretraining on diverse geospatial data may encode semantic knowledge of such landmarks, enabling the model to assign higher importance to this region in NLRS despite its proximity to the image edge. Across all three regions, the inclusion of AlphaEarth enables the model to better integrate spatial context across modalities and attend to semantically meaningful structures near the image boundary.

\section{Conclusion}
\label{sec:conclusion}

In this study, we have demonstrated an effective approach to incorporating geospatial foundation model embeddings into urban livability evaluation. We introduce a way to include geospatial foundation model embeddings through convolutions in a Transformer model. From this, we found that AlphaEarth and other embeddings can improve the performance of the original livability evaluation model and substantially mitigate performance degradation when individual modalities such as RS or POI are unavailable. However, AlphaEarth does not fully substitute for these modalities: when RS is excluded, the \textsc{aef} model still underperforms the full-data \textsc{aef} network, reflecting fine-grained spatial information present in high-resolution imagery that embeddings do not completely capture. 

Based on our results, we offer the following guidance for practitioners deploying this framework in new contexts. AlphaEarth consistently provides the most reliable performance gains and is a suitable embedding choice when operating under a single-embedding budget. In predominantly urban contexts, \textsc{aef} alone or \textsc{aef+tm} offers the best cost-benefit tradeoff. However, practitioners working in rural or mixed urban-rural regions should not rely on a single embedding — single embeddings like \textsc{aef} or \textsc{as} can worsen baseline performance in rural areas, and only multi-embedding combinations such as \textsc{aef+as+tm} reliably recover and improve upon the baseline. This likely reflects a distributional mismatch between the predominantly urban composition of embedding pretraining data and rural geospatial patterns, and should be treated as a deployment risk until embeddings with broader geographic coverage become available. When key modalities are unavailable, embeddings offer meaningful mitigation: \textsc{aef} alone recovers near full-data performance when DSM, NLRS, or POI are missing, and \textsc{aef+tm} approaches full-data baseline performance even without any of the four standard modalities. RS remains the hardest modality to substitute, as even the best embedding combinations do not fully close the performance gap when RS is absent.


We further examine interpretability through attention entropy analysis and find  that the higher \textsc{aef} attention weights observed in high-livability areas reflect greater input heterogeneity rather than increased reliance on  AlphaEarth specifically — high-livability areas distribute attention more broadly across modalities, while low-livability areas are dominated by POI.
We also plot Grad-CAM values and find that AlphaEarth embeddings can unify the Grad-CAM values of different modalities, identify important landmarks near the image edge, and increase the Grad-CAM values of NLRS.
While the embeddings do help generalizability, its performance boost is still lower than expected. This suggests that the embeddings are primarily helpful in geographical areas of data scarcity. However, there are no established validation techniques for rural livability evaluation. In future work, we plan to examine how embeddings contribute to spatial biases and how these can be mitigated through architecture design. Extending the framework beyond the Netherlands to test cross-cultural and cross-urban generalization is another important direction. Finally, incorporating temporal embeddings is an interesting research direction to explore how livability evolves over time, opening the door to longitudinal urban analysis.

\revdone{\textbf{Limitations.} All results in this paper are evaluated in the Netherlands, using the LBM dataset. Although we test on rural regions, we cannot yet validate whether these findings transfer to other regions that lack the ground-truth labels or the geospatial inputs necessary to train the model. We see this as a limitation and note it as an important area for future work. Applying this model to other livability or socioeconomic benchmarks outside the Netherlands would let us test whether the preference of AlphaEarth and multi-embedding combinations would hold on out-of-distribution areas that these embeddings were meant to be validated on.}

\begin{acks}
This research used the TGI RAILs advanced compute and data resource which is supported by the National Science Foundation (NSF) (award No. 2232860) and the Taylor Geospatial.
This material is based upon work supported by NSF under award No. 2118329 and by the NSF Graduate Research Fellowship Program under award No. DGE 2146756. Any opinions, findings, conclusions, or recommendations expressed in this material are those of the author(s) and do not necessarily reflect the views of NSF.
The authors acknowledge the use of Claude to develop code for the experiments and improve clarity and readability of the paper.
\end{acks}

\section*{Ethics and Privacy Statement}

This study uses only publicly available, non-personal data: satellite and elevation imagery, point-of-interest listings, geospatial foundation model embeddings, and the Dutch government's aggregated Leefbaarometer livability statistics, all reported at the level of 100~m $\times$ 100~m grid cells rather than individuals or households. No human-subjects data was collected, and no institutional review was required. The primary societal risk we identify is downstream misuse: livability predictions of this kind could be applied to real estate valuation, insurance underwriting, or resource-allocation decisions in ways that entrench existing inequality, particularly given the rural/urban performance disparities and embedding-driven biases we document. We intend this work as a research contribution to inform human-supervised policy and planning analysis, not as an input to automated eligibility or pricing decisions, and we report these disparities explicitly so that practitioners are aware of where the model is least reliable.

\bibliographystyle{ACM-Reference-Format}
\bibliography{sample-base}

\appendix

\section{Training Details}
\label{app:train}

All models were trained on a single NVIDIA H100 with 80 GB of HBM3 memory, utilizing CUDA 13.1 for GPU acceleration. The models were trained using PyTorch for 12 epochs with a batch size of 16. Our optimizer was AdamW~\cite{adamw} with a learning rate of $5e-5$, weight decay of $0.1$, epsilon of $1e-8$, and maximum gradient norm of $1$. We used early stopping with a patience of 5. We used BERT to create the text embedding from integer token indices with a sequence length of $512$. DenseNet extracted RS, DSM, and NLRS image embeddings of shape $3 \times 1024$. The convolution layers also extracted $3 \times 1024$ embeddings from AlphaEarth. The text and image inputs were concatenated and fed into a Transformer Encoder with 12 self-attention heads, 768 hidden units, and an activation function of Gaussian Error Linear Unit (GELU)~\cite{gelu}. We set Dropout to $0.1$, warmup steps to $0$, and gradient accumulation steps to $1$. Although the original paper~\cite{zhou_livability} utilized MSE as the loss function, we use MAE because it had slightly better performance on a majority of outputs, as illustrated in \Cref{tab:loss}. The table only recorded the performance for one model on the same seed.

\begin{table}[ht]
\centering
\caption{RMSE comparison between loss functions on baseline model across livability domain scores. Best values for each score are bolded.}
\label{tab:loss}
\begin{tabular}{lcccccc}
\toprule
\textbf{Model} & \textbf{LIV} & \textbf{PHY} & \textbf{NUI} & \textbf{SOC} & \textbf{AME} & \textbf{HOU} \\
\midrule
\textsc{mae}      & \bf 0.104 & \bf 0.026 & \bf 0.061 & \bf 0.031 & 0.054 & \bf 0.034 \\
\textsc{mse}      & \bf 0.104 & 0.027 & \bf 0.061 & \bf 0.032 & \bf 0.053 & \bf 0.034 \\
\bottomrule
\end{tabular}
\vspace{-0.5em}
\end{table}

\revdone{Each embedding source (AlphaEarth, AnySat, TerraMind) is processed by its own dedicated convolutional branch; branches do not share weights with each other or with the DenseNet/BERT backbones. Each branch takes the raw embedding as a $C_{\text{in}} \times H \times W$ tensor ($C_{\text{in}}=64$, $H{=}W{=}50$ for AlphaEarth; $C_{\text{in}}=1536$, $H{=}W{=}24$ for AnySat; $C_{\text{in}}=384$, $H{=}W{=}14$ for TerraMind) and applies three stacked $3\times3$ convolutions with stride 1 and padding 1, each mapping to 128 output channels and each followed by batch normalization and a ReLU activation. The resulting $128 \times H \times W$ feature map is spatially reduced by adaptive average pooling to $3 \times 1$ locations, flattened, and linearly projected from 128 to 1024 dimensions, followed by a ReLU and dropout with rate $0.1$, yielding the same $3 \times 1024$ token shape used by the DenseNet branches so all modalities can be concatenated before the Transformer Encoder.}

\section{\revdone{Comparison to Published TMTMR Baseline}}
\label{app:published}

\begin{table}[ht]
\centering
\caption{\revdone{RMSE comparison between our reimplemented \textsc{base} model and the published TMTMR baseline~\cite{zhou_livability} (\textsc{base-mse}) across livability domain scores. Best values for each score are bolded.}}
\label{tab:published}
\resizebox{1.0\linewidth}{!}{
\revdone{
\begin{tabular}{@{}lcccccc@{}}
\toprule
\textbf{Model} & \textbf{LIV} & \textbf{PHY} & \textbf{NUI} & \textbf{SOC} & \textbf{AME} & \textbf{HOU} \\
\midrule
\textsc{base-mse}~\cite{zhou_livability} & \textbf{0.0966} & \textbf{0.0248} & \textbf{0.0580} & \textbf{0.0294} & \textbf{0.0488} & \textbf{0.0327} \\
\textsc{base} & 0.106$\pm$0.002 & 0.027$\pm$0.000 & 0.061$\pm$0.001 & 0.032$\pm$0.001 & 0.055$\pm$0.002 & 0.034$\pm$0.000 \\
\bottomrule
\end{tabular}
}
}
\vspace{-0.5em}
\end{table}

\revdone{\Cref{tab:published} compares our reimplemented \textsc{base} model against the published TMTMR baseline from Ref.\cite{zhou_livability}, which we label \textsc{base-mse} since it is trained with an MSE loss. \textsc{base} consistently trails \textsc{base-mse} by roughly 5 to 11\% RMSE across all six scores. We attribute this gap to differences in training we did not attempt to replicate, such as \textsc{base-mse} being trained for 20 epochs instead of 12 or \textsc{base-mse} using MSE loss instead of MAE. We selected MAE because it performed marginally better in our own environment (\Cref{tab:loss}). Because of this gap, all embedding-related comparisons in this paper are made relative to our own \textsc{base} reimplementation rather than to \textsc{base-mse}, which we report here for reference only.}

\end{document}